\pdfoutput=1

\documentclass[11pt]{article}

\usepackage{acl}

\usepackage{times}
\usepackage{latexsym}

\usepackage[T1]{fontenc}

\usepackage[utf8]{inputenc}

\usepackage{microtype}

\usepackage{alphabeta}
\usepackage{tabularx}
\usepackage{booktabs}
\usepackage{amsmath}
\usepackage{amssymb}
\usepackage{amsfonts}
\usepackage{listings}
\usepackage{newfloat}
\usepackage[framemethod=TikZ]{mdframed}
\usepackage{multirow}
\usepackage{enumitem}
\usepackage{tikz}
\usepackage{textcomp}

\usepackage{xspace}
\def\cosi#1{#1} 
\def\base{\cosi{Gen}\xspace}
\def\dual{\cosi{DE}\xspace}
\def\adual{\cosi{AADE}\xspace}
\def\poly{\cosi{PE}\xspace}
\def\act{\cosi{AAE}\xspace}
\def\comb{\cosi{AARGH}\xspace}
\def\hialpha{$\uparrow$\xspace}
\def\loalpha{$\downarrow$\xspace}

\usepackage{array}
\newcommand{\PreserveBackslash}[1]{\let\temp=\\#1\let\\=\temp}
\newcolumntype{C}[1]{>{\PreserveBackslash\centering}p{#1}}
\newcolumntype{R}[1]{>{\PreserveBackslash\raggedleft}p{#1}}
\newcolumntype{L}[1]{>{\PreserveBackslash\raggedright}p{#1}}

\def\pz{\phantom{0}}
\usepackage[normalem]{ulem}

\title{Aargh! End-to-end Retrieve \& Generate for Task-oriented Dialog Systems }
\title{AARGH! End-to-end, Action-Aware Retrieval-Generation Hybrid for Task-oriented Dialog }
\title{AARGH! End-to-end, Action-Aware Retrieval-Generation Hybrid Model for Task-Oriented Dialog }
\title{AARGH! End-to-end Retrieve \& Generate for Task-Oriented Dialog }
\title{AARGH! End-to-end Retrieval-Generation for Task-Oriented Dialog }

\author{Tomáš Nekvinda \normalfont{and} \textbf{Ondřej Dušek} \\
  Charles University, Faculty of Mathematics and Physics \\
  Institute of Formal and Applied Linguistics \\
  Prague, Czech Republic \\
  \texttt{\{nekvinda,odusek\}@ufal.mff.cuni.cz} \\}

\date{}

\begin{document}
\maketitle
\begin{abstract}
We introduce \comb, an end-to-end task-oriented dialog system combining retrieval and generative approaches in a single model, aiming at improving dialog management and lexical diversity of outputs. The model features a new response selection method based on an action-aware training objective and a 
simplified single-encoder retrieval architecture which allow us to build an end-to-end retrieval-enhanced generation model where retrieval and generation share most of the parameters. 

On the 
MultiWOZ dataset, we show that our approach produces more diverse outputs while maintaining or improving state tracking and context-to-response generation performance, compared to state-of-the-art baselines.
\end{abstract}

\section{Introduction}

Most research task-oriented dialog models nowadays focus on end-to-end modeling, i.e., the whole dialog system is integrated into a single neural network \cite{wen2017,ham2020}. 
Although recent end-to-end \emph{generative} approaches based on pre-trained language models produce fluent and natural responses, they suffer from two major problems: (1) hallucinations and lack of grounding \citep{dziri2021}, which result in faulty dialog management or responses inconsistent with the dialog state or database results, and (2) blandness and low lexical diversity of outputs \citep{zhang2020dialogpt}. 
On the other hand, \emph{retrieval-based} dialog systems \citep{chaudhuri2018} 
select the most appropriate response candidate from a human-generated training set, thus producing varied outputs. However, their responses might not fit the context and can lead to disfluent conversations, especially when the set of candidates is sparse. This limits their usage to very large datasets which do not support dialog state tracking or database access \citep{lowe2015, alrfou2016}. 

Several recent works focus on combining the retrieval and generative dialog systems via response selection and subsequent refinement, i.e., \emph{retrieval-augmented generation} \citep{pandey2018, weston2018, cai2019a, thulke2021}. These models are used for open-domain conversations or to incorporate external knowledge into task-oriented systems and do not consider an explicit dialog state.

Our work follows the retrieve-and-refine approach, but we adapt it for database-aware task-oriented dialog. We aim at improving diversity of produced responses while preserving their appropriateness. In other words, we do not retrieve any new information from an external knowledge base, instead, we retrieve relevant training data responses to support the decoder in producing varied outputs.  To the best of our knowledge, we are the first to use retrieval-augmented models in this context. Unlike previous works, 
we merge the retrieval and generative components into a single neural network and train both tasks jointly, instead of using two separately trained models.
Our contributions are summarized as follows:\footnote{Code: \url{https://github.com/Tomiinek/Aargh}}

\begin{figure*}[!t]
\centering
\includegraphics[width=1.0\textwidth]{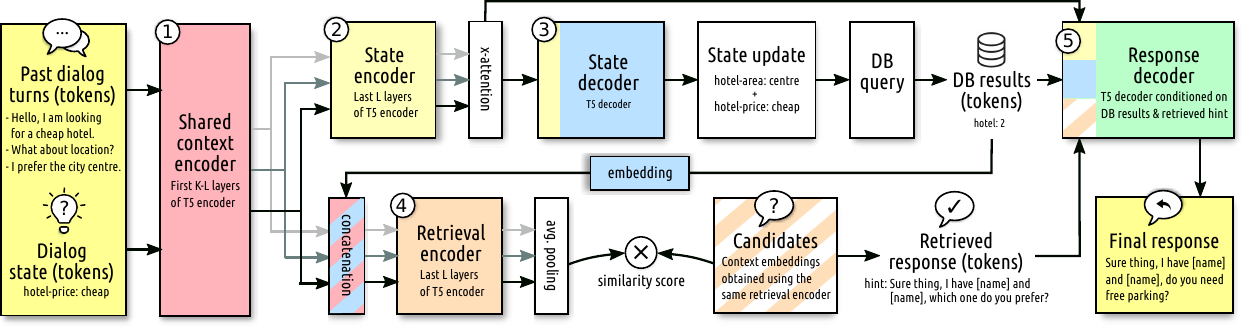}
\caption{Our retrieval-based generative task-oriented system (\comb, see Section~\ref{sec:hybrid}). Numbers in module boxes mark the order of processing during inference: (1) inputs are pushed through the shared context encoder and (2) state encoder; (3) the state decoder produces the update to the current dialog state. The new state is used to query the database whose outputs are discretized, embedded, and (4) used in the retrieval encoder whose output is reduced to a single vector via average pooling. The context embedding is used to get the best response candidate (hint). Finally, (5) the response decoder, which can attend to the state encoder outputs via cross-attention and is conditioned on the database results and the hint, generates the final system response to be shown to the user.}
\label{fig:pipeline}
\end{figure*}

\begin{itemize}[itemsep=0pt,topsep=4pt,leftmargin=12pt]

\item  We propose a single-encoder retrieval model utilizing dialog action annotation during training, and we show its superior retrieval capabilities in the task-oriented setting compared to two-encoder baseline models \cite{humeau2020}.

\item  We propose an end-to-end \emph{task-oriented} generative system with an \emph{integrated} minimalistic retrieval module. 
We compare it to strong
baselines that model response selection and generation separately. 

\item  On the MultiWOZ benchmark \citep{budzianowski2018large}, our approaches outperform previous methods in terms of lexical diversity and achieve competitive or better results in automatic metrics and human evaluation.
\end{itemize}

\section{Related Work}

\paragraph{Task-Oriented Response Generation} Most current works focus on building multi-domain data\-base-groun\-ded systems. The breeding ground for this research is the large-scale conversational dataset MultiWOZ \citep{budzianowski2018large, eric-etal-2020-multiwoz, zang2020}. 

Recent models often benefit from \emph{action annotation}. 
\citet{zhang2020damd} use action-based data augmentation and a three-stage architecture, decoding the dialog state, action, and response. \citet{chen2019} generate responses without state tracking, exploiting a hierarchical structure of the action annotation. 
On the other hand, reinforcement learning models 
\citep{wang2020} 
learn latent actions from data without using annotation.

Recent works focus on \emph{end-to-end} systems based on pre-trained language models. \citet{budzianowski2019} fine-tune GPT-2 \citep{radford2019} to model task-oriented dialogs, \citet{hosseini2020} enhance this approach 
with explicitly decoded system actions. \citet{peng2021} use auxiliary training objectives and machine teaching for GPT-2 fine-tuning. \citet{lin2020} introduced the encoder-decoder-based framework MinTL with BART \citep{devlin2019} or T5 \citep{kale2020} backbones (see Section~\ref{sec:baseline}).

\paragraph{Response Selection} can be viewed as scoring response candidates given a dialog context. A popular approach is the dual encoder architecture \citep{lowe2015, henderson2019a} where the response and context encoders model a joint embedding space. The encoders can take various forms: \citet{henderson2019b} compare encoders based on BERT \citep{devlin-etal-2019-bert} and custom encoders pre-trained on Reddit; \citet{wu2020} pre-train encoders specifically for task-oriented conversations. \citet{humeau2020} introduce poly-encoders, which produce multiple context encodings and add an attention layer to allow rich interaction with the candidate encoding (cf.\ Section~\ref{sec:selection}). 

\paragraph{Retrieval-Augmented Generation} 

To benefit from both retrieval and generative models, \citet{weston2018} proposed an open-domain dialog system utilizing a retrieval network and a decoder to refine retrieved responses. \citet{roller2021} further developed this approach, using poly-encoders with a large pre-trained decoder. They found that their decoder tends to ignore the retrieved response hints. To combat this, they propose the $\alpha$-blending method (replacing retrieval output with ground truth, see Section~\ref{sec:generation}).
Similarly, \citet{gupta2021} and \citet{cai2019b, cai2019a} focus on retrieval-augmented open-domain dialog, but to prevent the inflow of erroneous information into the generative part of their models, they use semantic frames or reduced forms of retrieved responses instead of raw response texts.

\citet{thulke2021} aim at knowledge retrieval from external 
documents for resolution of out-of-domain questions on MultiWOZ \citep{kim2020}. 
\citet{shalyminov2020} present the only work using generation and retrieval in a single model. They finetune GPT-2 \citep{radford2019} for response generation in a low-resource task-oriented setup, retrieve alternative responses based on the model's embedding similarity, and choose between generated and retrieved responses on-the-fly. However, their model is not trained for retrieval, cannot alter retrieved responses, and does not take a dialog state or database into account. 




\section{Method}
\label{sec:method}

We aim at end-to-end modeling of database-aware task-oriented systems, i.e., systems supporting both dialog state tracking and response generation tasks \cite{young2013}.
We combine retrieval and generative models to reduce hallucinations and boost output diversity.
We first describe our purely generative baseline (Section \ref{sec:baseline}), then explain baseline generation based on retrieved hints (Section \ref{sec:generation}). We then introduce baseline retrieval models (Section~\ref{sec:selection}) and our action-aware retrieval (Section~\ref{sec:action-aware}). Finally, we describe \comb, our single-model retrieval generation hybrid, in Section~\ref{sec:hybrid}.  \comb is shown in Figure~\ref{fig:pipeline}; other setups are depicted in Appendix~\ref{sec:appendix_a}. 

\subsection{Generative Baseline}
\label{sec:baseline}

Our purely-generative baseline model (\textbf{\base}) follows MinTL \citep{lin2020}. It is based on an encoder-decoder backbone with a context encoder, shared among two decoders: one for modeling the dialog state updates, the other for producing the final system response. Both decoders attend to the encoded input tokens via an attention mechanism.  

The encoder input sequence consists of a concatenation of two parts: (1) past dialog utterances prepended with \emph{<|system|>} or \emph{<|user|>} tokens, and (2) the initial dialog state converted to a string, e.g., \emph{hotel [area: center] restaurant [food: African, pricerange: expensive]}. The first decoder is conditioned only on the start-of-sequence token and predicts the dialog state update as a difference between the current state and the initial state. 
The second decoder is conditioned on the number of database results for each queried domain, e.g. \emph{train: 6} if there are six matching results for a train search, and generates the final response.

During inference, the input is passed through the encoder, then the state update is predicted, merged with the initial dialog state, and this new 
state is used to query the database (see Section~\ref{sec:experimental_setup} for details). 
The final system response is predicted based on the context, state, and database results.

\subsection{Retrieval-Augmented Response Generation}
\label{sec:generation}

To combine the retrieval and generative approaches, we follow \citet{weston2018} and incorporate response \emph{hints}, i.e., the outputs of a retrieval module (Sections~\ref{sec:selection}, \ref{sec:action-aware}), into the generative module in their original form as raw sub-word tokens. Specifically, we add the retrieved response prepended with \emph{<|hint|>} to the input of \base's response decoder (Section~\ref{sec:baseline}), alongside the database results.

\citet{gupta2021} state that this straightforward token-based retrieve \& refine setup might lead to generating incoherent responses due to over-copying of contextually irrelevant tokens. However, using more abstract outputs of the retrieval module, e.g.\ semantic frames or salient words would go against our goal of reducing blandness and increasing responses lexical diversity. To smoothly control the amount of token copying, we follow \citet{roller2021} and use the so-called $\alpha$-blending. During training, we replace the retrieved utterance with the ground-truth final response with probability $\alpha$. This method also ensures that the decoder learns to attend to the retrieval part of its input successfully.


\subsection{Baseline Response Selection}
\label{sec:selection}

We consider two baseline retrieval model variants: 

\paragraph{Dual-encoder (\dual)} follows the very popular retrieval architecture \citep{lowe2015,humeau2020} which makes use of context and response encoders. Both produce a single vector in a joint embedding space. During training, the context embedding and the corresponding response embedding are pushed towards each other, while other responses in the training batch are used as negative examples, i.e., cross-entropy loss is used:
$$ \mathcal{L}(S) = \frac{1}{N} \sum_{j}{\left( -S_{j, j} + \log{\sum_{i}{e^{S_{j, i}}}} \right)} $$
where $S \in \mathbb{R}^{N \times N}$ is the similarity matrix between \emph{normalized} encoded responses $\mathbf{e}^r$ and contexts $\mathbf{e}^c$ in the batch, specifically $S_{i,j} = w \cdot (\mathbf{e}^c_{i} \cdot \mathbf{e}^r_{j})$, where $w > 0$ is a trainable scaling factor. 

Inference-time retrieval is as simple as finding the nearest candidate embedding given a context embedding.  The context input is similar to \base's (see Section~\ref{sec:baseline}): a concatenation of the current updated dialog state, the number of matching database results and past user and system utterances. 
Encoders are followed by average pooling and a fully-connected layer for dimensionality reduction.

\paragraph{Poly-encoder (\poly)} an extension of \dual, aiming at richer interaction between the candidate and the context. The candidate encoder is unchanged. In the context encoder, the average pooling is replaced with two levels of dot-product attention \citep{vaswani2017, humeau2020}. The first level summarizes the encoded context tokens into $m$ vectors. The context tokens act as attention keys and values; queries to this attention are $m$ learned embeddings (query codes). The second attention level provides the candidate-context interaction: it takes the $m$ context summary vectors as keys and values, and the candidate encoder output acts as the query. The parameter $m$ provides trade-off between inference complexity and richness of the context encoding. The loss term remains the same.

\subsection{Action-aware Response Selection}
\label{sec:action-aware}

We argue that the dual- or poly-encoder models are not practical for the 
task-oriented settings as their performance depends on the way negative examples are sampled during training \citep{nugmanova2019}.
Choosing appropriate negative examples is difficult in task-oriented datasets as system responses are often very similar to each other (with the conversations being in a narrow domain and following similar patterns).
Therefore, we propose a 
method for candidate selection based on system action annotation, which is usually available in 
task-oriented datasets. 
We designed the method to be usable with a single encoder only, but we also include a dual-encoder version for comparison.

\paragraph{Action-aware-encoder (\act)}  

Using two separate encoders to encode the response and the context might be impractical due to large model size. Some recent works (e.g., \citet{wu2020, roller2021}) use a single shared encoder instead, and \citet{henderson2020} discuss parameter sharing between the two encoders. 
In view of that, we propose a single-encoder action-aware retrieval model. We train it to produce embeddings of dialog contexts which are close to each other if the corresponding responses in the training data have similar action annotation. 
More precisely, we adapt \citet{wan2018}'s generalized end-to-end loss, originally developed for batch-wise training of speaker classification from audio: 
To form training mini-batches, we first sample $M$ random dialog actions, and for each of those actions, we sample $N$ examples that include the particular action in their system action annotation. We then encode dialog contexts corresponding to the sampled examples into normalized embeddings $\mathbf{e}_{m, n}$, and compute the similarity matrix as follows: 
\begin{equation*}
  S_{ji, k} =
    \begin{cases}
      w \cdot (\mathbf{e}_{j, i} \cdot \mathbf{c}^{\{i\}}_j) & \text{if k = j}\\
      w \cdot (\mathbf{e}_{j, i} \cdot \mathbf{c}^{\varnothing}_k) & \text{otherwise}
    \end{cases}  
\end{equation*}
\begin{equation*}
  \mathbf{c}^A_j = 
    \frac{1}{N - |A|} \sum_{i \in [N] - A} \mathbf{e}_{j,i}
\end{equation*}
where $S \in \mathbb{R}^{N \cdot M \times M}$, $i \in [N]$, $j,k \in [M]$, and $A \subseteq [N]$ is a set of indices. Same as for \dual, $w > 0$ is a trainable scaling factor of the similarity matrix. In other words, the similarity matrix describes the similarity between embeddings of each example and centroids, i.e., the means of $N$ embeddings that correspond to the same particular action. For stability reasons and to avoid trivial solutions, we follow \citet{wan2018} and exclude $\mathbf{e}_{j, i}$ from the centroid calculation when computing $S_{ji, j}$.

We then maximize the similarity between the examples and their corresponding centroids while using other centroids as negative examples:
$$ \mathcal{L}(S) = \frac{1}{N \cdot M} \sum_{j,i}{\left( -S_{ji, j} + \log{\sum_{k}{e^{S_{ji, k}}}} \right)} $$
During inference, we rank the responses from the training set according to the cosine similarity of their corresponding contexts and the query context. Again, the contexts consist of the current updated dialog state, the number of matching database results and past utterances. 


\paragraph{Action-aware-dual-encoder (\adual)} 
This setup follows the \dual architecture (see Section~\ref{sec:selection}), but it is trained in a similar way as \act, i.e., we form training mini-batches identically and for each of $M$ distinct actions in the batch, we treat all $N$ examples as positive examples. 

\subsection{Hybrid End-to-end Model}
\label{sec:hybrid}

To further simplify the retrieval-augmented setup, reduce the number of trainable parameters and gain back computational efficiency, 
we introduce an end-to-end Action-Aware Retrieval-Generative Hybrid model (\textbf{\comb}), which jointly models both response selection and context-to-response generation (see Figure~\ref{fig:pipeline}). It is a natural extension of the \base generative model (Section~\ref{sec:baseline}), enabled by our new single-encoder action-aware response retrieval (\act, Section~\ref{sec:action-aware}).  

A new retrieval encoder, which produces normalized context embeddings, shares most parameters with the original 
encoder, which is followed by the two decoders and is partially responsible for state tracking and response generation.
To build the retrieval encoder, we fork the last $L$ layers of the original
encoder and condition them on the outputs of the shared preceding layers, concatenated with an embedding of the number of current database results. To obtain this embedding, 
we convert the number of database results into a small set of bins, which are then embedded via a learnt embedding layer of size $E$.\footnote{This conversion is dataset-specific and not used in other compared models such as \base. We use the label 0 if there are no results, 1 for 0 matching results, 2, 3, 4 if there are 1, 2 or 3 results, respectively, 5, 6 if there are less than 6 or 11 results, and 7 if there are 11 or more results.} 
The new retrieval  encoder is followed by average pooling and trained using the same objective 
as \act (see Section~\ref{sec:action-aware}). 

During inference, we pass the input through the partially shared context encoder and decode and update the dialog state. The new state is used to query the database. Database results are embedded and added to the output of the last encoder shared layer to form the input to the retrieval encoder, which produces the context embedding and a retrieved response. Based on state, database results, and retrieved response, the response decoder produces the final (delexicalized) response.


\section{Experimental Setup}
\label{sec:experimental_setup}

\paragraph{Models} Our models are based on pre-trained models from HuggingFace \citep{wolf-etal-2020-transformers}: We implement \base and the generative parts in our retrieval-based models using T5-base \citep{kale2020}.
Retrieval encoders in \dual, \adual, \poly and \act are implemented as fine-tuned BERT-base \citep{devlin2019}. \comb is built upon T5-base, same as \base; we fork the last $L=2$ out of $K=12$ encoder layers. The choice of $L$ is a trade-off between model performance and size.\footnote{We noticed a performance drop when using $L=1$, and $L=3$ did not bring any large gain.} The database embedding has size $E=4$. For simplicity, we do not use specialized backbones pre-trained on dialogs 
such as ToD-BERT \citep{wu2020}. \poly uses $m=16$ query codes (see Section~\ref{sec:selection}) and single-headed attention mechanisms.

\begin{figure}[!t]
\centering
\includegraphics[width=0.48\textwidth]{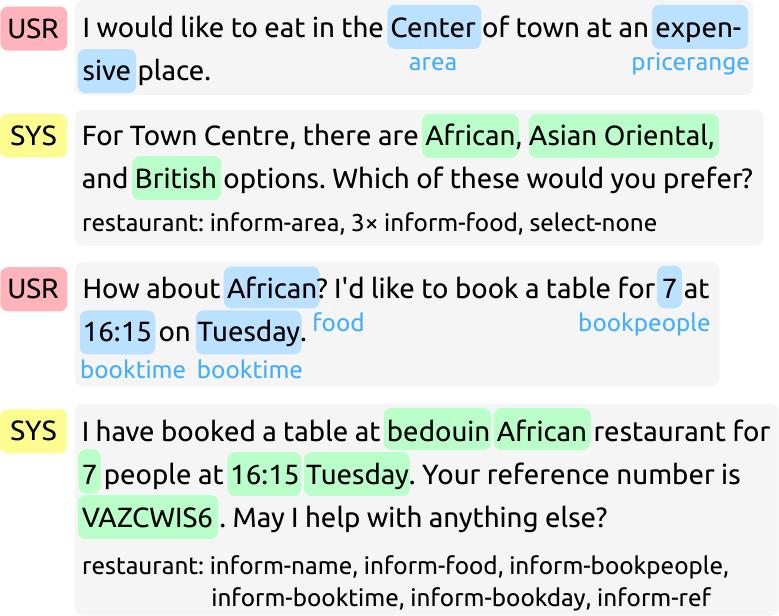}
\caption{Part of a short conversation from MultiWOZ. It has
\definecolor{r}{RGB}{255,179,186} \tikz\draw[r, fill=r] (0,0) circle (.7ex); user and 
\definecolor{yy}{RGB}{252,252,144} \tikz\draw[yy, fill=yy] (0,0) circle (.7ex); system turns, and annotated \definecolor{g}{RGB}{186,255,200} \tikz\draw[g, fill=g] (0,0) circle (.7ex); slot spans.
Both, user and system affect the
\definecolor{b}{RGB}{186,225,255} \tikz\draw[b, fill=b] (0,0) circle (.7ex); dialog state. Actions are shown below system texts.
}
\label{fig:example}
\end{figure}

\begin{figure*}[!t]
\centering
\includegraphics[width=1.0\textwidth]{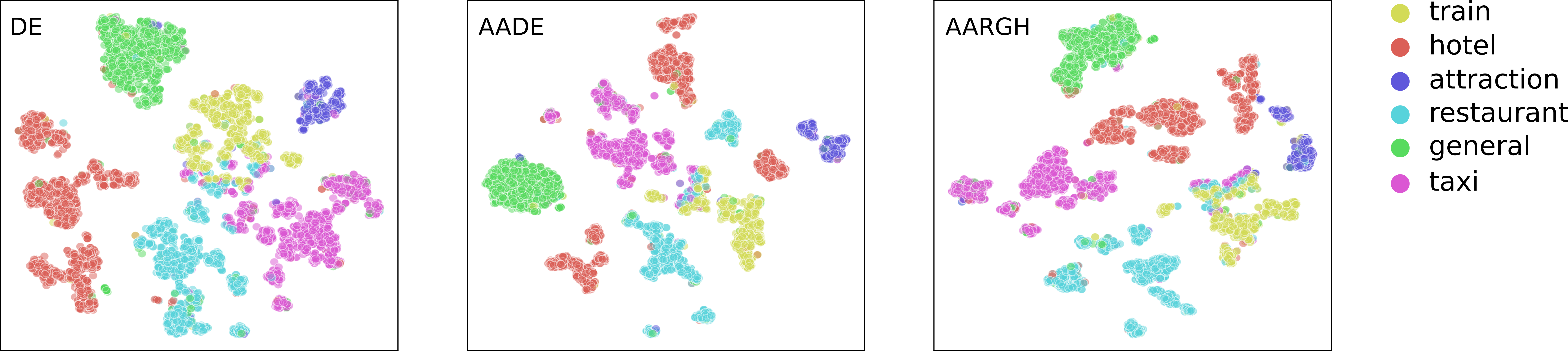}
\caption{t-SNE projection of test set context embeddings (colored by domains) of retrieval modules of our models. The colors indicate the different MultiWOZ domains that are associated with the corresponding dialog turns.
}
\label{fig:clustering}
\end{figure*}

\paragraph{Data and database} We experiment on the MultiWOZ~2.2 dataset \citep{budzianowski2018large, zang2020} which is a popular dataset with around 10k task-oriented conversations in 7 different domains such as trains, restaurants, or hotels (see Figure~\ref{fig:example}). A single conversation can touch multiple domains. The dataset has an associated database, dialog state annotation, dialog action annotation of system turns, and slot value span annotation for easy delexicalization \citep{wen2015}, thus enabling development of realistic end-to-end dialog systems.\footnote{Unlike the similar-sized Taskmaster \citep{byrne2019} and SGD \citep{rastogi2019} datasets, which lack databases and annotation detail.} To query the database using the belief state, we use the fuzzy matching implementation by \citet{nekvinda2021}. To filter out inactive domains from database results during inference, we follow previous work and estimate the currently active domain from dialog state updates.


\paragraph{Input and output format} We use the same formats for all models.
Target responses are delexicalized using MultiWOZ 2.2 span annotation, and we limit the context to 5 utterances. MultiWOZ action labels include domain, action, and slot name, e.g., \emph{train-inform-price}. 
We remove domains from the labels to limit data sparsity.

\paragraph{Training procedure} \dual, \adual, \poly and \act are trained in two stages. The retrieval part is trained first and provides response hints to the generative model during the second phase. Modules in \comb are trained jointly, but we alternate parameter updates for the retrieval encoder and the rest of the network. To do so, we use two separate optimizers. \comb's hints used in the response decoder during training are refreshed after every epoch. All models are optimized using Adam \citep{kingma2014} and cosine learning rate decay with warmup. With respect to memory limits of our hardware, we set $N=6$, $M = 8$ for batch sampling during training of retrieval parts of \act and \comb.

\paragraph{$\mathbf{\alpha}$-blending} We experiment with two $\alpha$-blending values: a conservative one ($\alpha=0.05$, marked “\loalpha”) and a greedy one ($\alpha=0.4$, marked “\hialpha”), targeting a mostly generation-focused and a mostly retrieval-focused setting.\footnote{The values were chosen empirically, based on preliminary experiments on development data.}

\paragraph{Decoding} We use greedy decoding for dialog state update generation. For response generation, we report results with greedy decoding in Section~\ref{sec:results} and with beam search in Appendix~\ref{sec:appendix_b}.

\section{Evaluation and Results}
\label{sec:results}

We focus on end-to-end modeling, which includes dialog state tracking and response generation. All reported results are on MultiWOZ test set with 1000 dialogs, averaged over 8 different random seeds. We generated responses given ground truth contexts. We follow MinTL and predict the dialog state cumulatively for each conversation turn, which means that state tracking errors may compound.
See Appendix~\ref{sec:appendix_c} for an example end-to-end conversation without any ground-truth information.

\begin{table}[t!]
    \centering\small
    \setlength{\tabcolsep}{3.0pt}
    \begin{tabular}{l|cc >{\centering}m{0.115\linewidth} >{\centering}m{0.115\linewidth} >{\centering\arraybackslash}m{0.12\linewidth}}
      \toprule
      Setting    & BLEU & 
      Action IoU & \% full match & 
      \% no match   & \% uniq. hints \\
      
      \midrule
            Random & \pz2.1  &  \pz5.1  \textpm~0.2  & \pz1.1  & 85.0 & 93.5 \\
            \dual  & \pz8.9  &  34.7 \textpm~0.5  & 11.0 & 29.7 & 54.1 \\
            \adual & \pz7.9  &  30.9 \textpm~1.7  & \pz8.9  & 33.4 & 24.2 \\
            \poly  & \pz8.8  &  35.0 \textpm~0.8  & 11.4 & 28.9 & 44.1 \\
            \act   & \textbf{12.8} &  \textbf{37.1} \textpm~0.2  & \textbf{14.5} & \textbf{28.6} & 88.6 \\
            \comb  & 12.6 &  36.6 \textpm~0.2  & 14.2 & 29.0 & \textbf{89.6} \\
      
      \bottomrule
    \end{tabular}
    \caption{Evaluation of retrieval components of our models (Section~\ref{sec:selection}, \ref{sec:hybrid}). See Section~\ref{sec:response_results} for details.}
    \label{tab:retrieval_eval}
\end{table}

\subsection{Response selection}
\label{sec:response_results}

\begin{table*}[t!]
    \centering\small
    \setlength{\tabcolsep}{2.15pt}
    \begin{tabular}{cl|c>{\hspace{1mm}}c>{\hspace{1mm}}ccc|cc|c}
      \toprule
      \multicolumn{2}{c|}{Setting} & BLEU & Inform & Success & Unique trigrams & BCE & Hint-BLEU & Hint-copy & Joint acc. \\
      
      \midrule
      
      \multicolumn{2}{l|}{Corpus}        & -      & 93.7 & 90.9 & 25,212 & 3.37 & - & - & - \\
      \multicolumn{2}{l|}{SOLOIST \citep{peng2020}}      & 13.6 & 82.3 & 72.4 & \pz7,923  & 2.41 & - & - & - \\
      \multicolumn{2}{l|}{PPTOD \citep{pttod2021}}       & 18.2 & 83.1 & 72.7 & \pz2,538  & 1.88 & - & - & - \\
      \multicolumn{2}{l|}{MTTOD \citep{mttod2021}}       & 19.0 & 85.9 & 76.5 & \pz4,066  & 1.93 & - & - & - \\
      \multicolumn{2}{l|}{MinTL \citep{lin2020}}         & 19.4 & 73.7 & 65.4 & \pz2,525  & 1.81 & - & - & - \\
      
      \midrule
      
      \multicolumn{2}{l|}{\base (equiv.~to MinTL)}         & 18.6 \textpm~0.3 & 77.0 \textpm~1.2 & 66.4 \textpm~1.0 & \pz3,209  & 1.94 & - & - & 54.1 \textpm~0.2 \\
      \multicolumn{2}{l|}{\act (retrieval only)} & 12.8 \textpm~0.1 & 79.9 \textpm~0.6 & 58.3 \textpm~0.7 & 22,457 & 3.34 & 100.0 & 100.0 \% & - \\
      
      \midrule
      
      \parbox[t]{2mm}{\multirow{5}{*}{\rotatebox[origin=c]{90}{\emph{$\alpha$=0.05}}}} 
            &  \dual+\base\loalpha & \textbf{17.6} \textpm~0.3 & 80.9 \textpm~0.5 & 68.8 \textpm~0.6 & \pz8,190 & \textbf{2.36} & 32.5 & 15.2 \% & 54.2 \textpm~0.1 \\
            & \adual+\base\loalpha & 17.3 \textpm~0.3 & 81.2 \textpm~0.9 & 69.1 \textpm~1.0 & \pz6,613 & 2.29 & 26.7 & 12.8 \% & 54.3 \textpm~0.1 \\
            &  \poly+\base\loalpha & 17.4 \textpm~0.3 & 79.9 \textpm~0.9 & 66.8 \textpm~1.0 & \pz7,736 & 2.35 & 31.3 & 14.5 \% & \textbf{54.4} \textpm~0.2 \\
            &  \act+\base\loalpha  & 17.5 \textpm~0.6 & \textbf{82.0} \textpm~1.0 & \textbf{70.3} \textpm~0.8 & \pz8,152 & 2.32 & 32.0 & 16.2 \% & 54.2 \textpm~0.2 \\
            &  \comb\loalpha & 17.3 \textpm~0.3 & 81.2 \textpm~0.6 & 69.5 \textpm~0.5 & \textbf{\pz8,200} & 2.33 & 28.4 & 14.2 \%  & 53.8 \textpm~0.2 \\
      
      \midrule
      
      \parbox[t]{2mm}{\multirow{5}{*}{\rotatebox[origin=c]{90}{\emph{$\alpha$=0.4}}}} 
        &  \dual+\base\hialpha & 12.3 \textpm~0.3 & 87.8 \textpm~0.3 & 69.1 \textpm~0.5 & 18,800 & 3.20 & 80.4 & 76.5 \% & 54.2 \textpm~0.2 \\
        & \adual+\base\hialpha & \textbf{14.6} \textpm~0.4 & 81.0 \textpm~0.8 & 66.7 \textpm~0.4 & 10,723 & 2.72 & 51.7 & 44.8 \% & 54.2 \textpm~0.1 \\
        &  \poly+\base\hialpha & 12.9 \textpm~0.4 & 86.0 \textpm~0.8 & 67.1 \textpm~0.6 & 16,632 & 3.13 & 74.0 & 69.1\% & \textbf{54.4} \textpm~0.1 \\
        &  \act+\base\hialpha  & 11.9 \textpm~0.2 & \textbf{90.5} \textpm~0.3 & \textbf{71.3} \textpm~0.3 & 19,436 & \textbf{3.23} & 91.1 & 89.3 \% & 54.3 \textpm~0.2 \\
        &  \comb\hialpha & 12.1 \textpm~0.2 & 89.6 \textpm~0.2 & 70.7 \textpm~0.5 & \textbf{19,813} & 3.21 & 87.6 & 85.0 \% & 53.6 \textpm~0.2 \\
     
      \bottomrule
    \end{tabular}
    \caption{Response generation and state tracking evaluation on MultiWOZ using automatic metrics, including the bi-gram conditional entropy (BCE) and number of unique trigrams. We compare previous work, the baseline and retrieval-based generative models. See Section~\ref{sec:automatic} for details about the metrics; Section~\ref{sec:method},~\ref{sec:experimental_setup} for model descriptions.
    }
    \label{tab:automatic_eval_greedy}
\end{table*}

First, we assess the performance of retrieval components of \dual, \adual, \poly, \act and \comb. We cannot use the popular R@k metric \citep{chaudhuri2018} as \act and \comb 
use embeddings of dialog contexts (not responses) of candidates as the search criterion
and would always score 100\%. 
Instead, we use the action annotation and measure the intersection over union (IoU), full-match and no-match rates on sets of actions associated with top-1 retrieved and ground-truth responses. 
We add BLEU \citep{papineni2002, liu2016} between ground-truth and retrieved responses and the proportion of distinct retrieval outputs to assess their lexical similarity to references and diversity. 

Table~\ref{tab:retrieval_eval} shows that \act and \comb significantly outperform other setups on all measures except for the no-match rate,\footnote{According to a paired t-test with 95\% confidence level.} where \poly has comparable results. This is expected as they use the additional action annotation during training, unlike \dual and \poly. \adual performs surprisingly bad. According to the unique hints rate, \act and \comb retrieve a much wider range of outputs, which could improve lexical diversity of final responses. The higher BLEU, Action IoU and full match rates suggest that the models retrieve responses more similar to the ground truth. 

\begin{table}[t!]
    \centering\small
    \setlength{\tabcolsep}{3.8pt}
    \begin{tabular}{l|cccc}
      \toprule
      Silhouette coefficient & \dual & \adual & \act & \comb \\
      
      \midrule
      per Domain & 0.098 & \textbf{0.179} & 0.151 & 0.159 \\
      
      per Action & 0.147 & 0.316 & 0.312 & \textbf{0.320} \\
      
      \bottomrule
    \end{tabular}
    \caption{Evaluation of domain and action separation (Section~\ref{sec:response_results}). We show averages over 8 random seeds.}
    \label{tab:clustering_eval}
\end{table}

To further compare the approaches to response selection, we computed the Silhouette coefficient \citep{rousseeuw1987} based on the active domain and action annotation (see Table~\ref{tab:clustering_eval}).\footnote{In the case of action-based clustering, we treat each action as a separate cluster; each example can belong to multiple clusters. The clustering measure is calculated for each cluster and averaged over all actions which are weighted by the size of the corresponding clusters.} We omit \poly because 
its context embeddings depend on queries, i.e., the candidate embeddings (other models output the same context regardless of candidates). 
\dual has the worst results; other systems perform similarly, but \comb is the best on action separation while \adual has the best scores for domains.

We see that \adual's context encoder is successful in clustering, but it lags behind in terms of correct action selection. Unlike \comb and \act, \adual retrieves candidates based on response embeddings. We hypothesize that lower response variability (compared to context variability) leads the model to prefer 
responses seen more frequently during training. 
\comb and \act are not affected by this as they use purely context-based retrieval. 

Figure~\ref{fig:clustering} provides a visualisation of the domain clusters projected using t-SNE \citep{vandermaaten2008}. It supports the findings of our evaluation based on the Silhouette coefficient: We see that visualisations of \comb and \adual embedding spaces look similarly whereas \dual's clusters appear more noisy.


\subsection{Response generation}

\begin{table*}[t]
    \centering\small
    \setlength{\tabcolsep}{4pt}
    \begin{tabular}{p{0.09\linewidth} p{0.875\linewidth}}
      \toprule
      \textbf{System} & \textbf{Response} \\
      
      \midrule
      \multicolumn{1}{l}{\textbf{Context}} & \begin{tabular}[t]{@{}p{\linewidth}@{}} 
      User: I need a taxi to take me to Yippee Noodle Bar by 9:45.\end{tabular} \\
      \addlinespace[0.1cm]
      
      Corpus   & Absolutely, where will the taxi be picking you up from today? \\
      \base    & I can help with that. Where are you departing from? \\
      \dual+\base\loalpha    & Where will you be departing from? \\
      \comb\loalpha   & Okay, I can help with that. Where are you coming from? \\
      \comb\hialpha   & Of course, I just need to know your \textbf{destination} so I can book that for you. \\
      
      \midrule
      \multicolumn{1}{l}{\textbf{Context}} & \begin{tabular}[t]{@{}p{\linewidth}@{}} 
      [...] \\
      User: It does, thanks. Can you tell me about a hotel called the Warkworth House? \\ 
      System: I sure can. They are a guesthouse in the east. They are moderately priced and 4-star rated. They also \\
      \quad\quad offer free internet and free parking. Can I help book a room? \\ 
      User: Yes I would like to book it for 3 nights starting on Wednesday for 4. \end{tabular} \\
      \addlinespace[0.1cm]
      
      Corpus   & I have made the reservation for you. The confirmation number is [ref]. They will expect you on [bookday]. Is there anything else I can get for you? \\
      \base    & I have made those reservations and your reference number is [ref]. \\
      \dual+\base\loalpha    & I'd be happy to make that booking for you. For how many people? \\
      \comb\loalpha   & I have successfully booked a room for you \textbf{at [name] for [bookpeople] starting on [bookday] for [bookstay] nights}. Your reference number is [ref]. Would that be all for you? \\
      \comb\hialpha   & You are all set! Reference number [ref]. Is there anything else I can help you with? \\
      
      \midrule
      \multicolumn{1}{l}{\textbf{Context}} & \begin{tabular}[t]{@{}p{\linewidth}@{}} User: I am planning a trip in Cambridge \\
      \end{tabular} \\
      \addlinespace[0.1cm]
      
      Corpus   & Great! Is there a certain restaurant you're looking for? \\
      \base    & I can help you with that. What type of food are you interested in? \\
      \dual+\base\loalpha    & I can help you with that. What type of food are you looking for? \\
      \comb\loalpha   & Great! Do you need assistance with \textbf{transportation arrangements}? \\
      \comb\hialpha   & Great! Would you like to book a train to take you there? \\
      
       \bottomrule
    \end{tabular}
    \caption{Side-by-side system output comparison (see Sections~\ref{sec:method} and~\ref{sec:experimental_setup} for model details, see Section~\ref{sec:human-eval} for example description). 
    The responses are continuations of conversations SNG0016, MUL1091, and PMUL4542 from MultiWOZ.
    }
    \label{tab:side-by-side2}
\end{table*}

We evaluate the response generation abilities of our models using automatic metrics and human assessment of delexicalized texts (see Table~\ref{tab:side-by-side2} for examples).

\paragraph{Evaluation with automatic metrics}
\label{sec:automatic}

We use the corpus-based evaluator by \citet{nekvinda2021} to measure commonly used metrics on MultiWOZ (Inform \& Success rates, BLEU) as well as lexical diversity measures, namely the number of distinct trigrams in the outputs and bigram conditional entropy \cite{li_diversity-promoting_2016,novikova_lexical_2019}. State tracking joint accuracy is calculated with scripts adapted from TRADE \citep{trade2019}.
To better understand the effect of using retrieved hints and to quantify the amount of copying, we calculate BLEU between retrieved hints and final generated responses (Hint-BLEU) and the proportion of generated responses exactly matching the corresponding retrieved hints (Hint-copy).

We include comparisons with recent strong end-to-end models on MultiWOZ: 
SOLOIST \citep{peng2020}, MTTOD \citep{mttod2021}, PPTOD \citep{pttod2021}, and MinTL \citep{lin2020}, which has the same architecture as \base. To show the importance of the generative parts of our models, we also include \act without the refining decoder. 

Table~\ref{tab:automatic_eval_greedy} shows scores obtained with greedy decoding (see Appendix~\ref{sec:appendix_b} for beam search results). All models have similar state tracking performance. \comb has slightly lower numbers, which is not surprising as it shares a substantial part of the encoder with its retrieval component. As expected, we notice a huge difference in Hint-BLEU and Hint-copy of versions with different $\alpha$-blending probabilities (\loalpha vs.~\hialpha).\footnote{Hint-copy of 15\% roughly means one turn per dialog.}
The performance boost over \base and retrieval-only \act is, for \loalpha variants, mainly in terms of Success. In \hialpha, more frequent hint copying reduces BLEU and improves lexical diversity; we also see higher Inform. \act+\base and \comb (both \loalpha and \hialpha) perform better than corresponding \dual+\base or \poly+\base on Inform and Success rates.\footnote{According to a paired t-test with 95\% confidence level.} Differences between \act+\base and \comb are not statistically significant and their Success scores are better than MinTL,  competitive with PPTOD and SOLOIST but lower than MTTOD. In terms of lexical diversity, all models are better than most generative baselines.\footnote{The \loalpha variants are similar to SOLOIST, which, however, reaches diversity by employng sampling \citep{holtzman2020} instead of greedy decoding.}

\begin{table}[t]
    \centering\small
    \setlength{\tabcolsep}{1.25pt}
    \begin{tabular}{lcccc}
      \toprule
              & \base & \dual+\base\loalpha & \comb\loalpha & \comb\hialpha \\
     \midrule
      
      Mean Ranking  & 2.03 & 1.99 & \textbf{1.91} & 2.3 \\
      Ranked \#1    & 36.1\% & 35.5\% & \textbf{40.8\%} & 37.9\% \\
      Ranked \#2    & 34.7\% & 36.7\% & 33.5\% & 18.8\% \\
      Ranked \#3    & 18.8\% & 20.5\% & 19.7\% & 18.8\% \\
      Ranked \#4    & 10.4\% & \pz7.2\%  & \pz\textbf{6.1\%}  & 24.6\% \\
      
      \bottomrule
    \end{tabular}
    \caption{Human evaluation results – mean ranks (1-4) established from 50 evaluated conversations.}
    \label{tab:human}
\end{table}

\paragraph{Human evaluation}
\label{sec:human-eval}

We arranged an in-house human evaluation on the delexicalized outputs of \base (i.e., MinTL's architecture), \dual+\base\loalpha, \comb\loalpha and \comb\hialpha. We used side-by-side relative ranking evaluation, which has been repeatedly found to increase consistency compared to rating isolated examples \citep{callison2007,belz_comparing_2010,kiritchenko_best-worst_2017}. Participants were given full dialog context and current database results, and we asked them to rank responses of the compared models from the best-fitting to the worst, where multiple responses could be ranked the same (see Appendix~\ref{sec:appendix_d} for details). We collected rankings for 346 turns of 50 conversations from 5 linguists with experience in natural language generation. All of them were given a different set of dialogs and they were instructed to focus on consistency with the context and database results, naturalness, and attractiveness of the responses. See Table~\ref{tab:human} for results.

Although \comb\hialpha scored the best on automatic metrics, it has worse mean ranks than other models, which all have similar mean ranks.\footnote{According to Friedman test with 95\% confidence level and Nemenyi post-hoc test; only the difference between \comb\hialpha and other models is statistically significant.} 
This confirms previous findings of low correlation between automatic metrics and human assessments \cite{liu_how_2016,novikova_why_2017}.
Upon detailed manual error analysis, we found that \comb\hialpha often copies whole hints including words that do not fit the context (see the top example in Table~\ref{tab:side-by-side2}), i.e., contradictions to earlier statements or noisy non-delexicalized values from the training set. \mbox{\comb\loalpha} performs slightly better than the baselines and is more often ranked best and least often ranked worst. The bottom two examples in Table~\ref{tab:side-by-side2} demonstrate that \comb\loalpha is often able to provide more detailed responses and uses more varied vocabulary, compared to the baseline.


\section{Conclusion}

We present \comb, an end-to-end task-oriented dialog system, combining retrieval and generative approaches. It uses an embedded single-encoder retrieval component which extends a purely generative model without the need for a large number of new parameters. \comb features an action-aware response selection training objective. Our experiments on the MultiWOZ dataset show that \comb outperforms baselines in terms of automatic metrics and human evaluation and it is competitive with state-of-the-art models such as SOLOIST or MTTOD. We showed that our proposed action-aware retrieval training objective supports retrieval of a larger variety of unique and relevant responses in the task-oriented setting and makes efficient use of the available system action annotation. Further, using the retrieval module improves dialog management in terms of the Success rate. A limitation of our approach is the need for careful hyper-parameter setting, coupled with the risk of overuse of retrieved responses that match the dialogue state but are not appropriate for the context. 

In future work, we would like to confirm our results on more datasets and explore more complex ways of usage of the retrieved responses to encourage the model to copy interesting language structures while ignoring inappropriate tokens or relics of faulty delexicalization. 

\section*{Acknowledgements}

This research was supported by Charles University projects GAUK 373921, SVV 260575 and PRIMUS/19/SCI/10, and by the European Research Council (Grant agreement No. 101039303 NG-NLG). 
It used resources provided by the LINDAT/CLARIAH-CZ Research Infrastructure (Czech Ministry of Education, Youth and Sports project No. LM2018101).

\bibliography{acl}
\bibliographystyle{acl_natbib}

\appendix

\section{Model Architectures}
\label{sec:appendix_a}

\begin{figure*}[!th]
\centering
\includegraphics[width=1.0\textwidth]{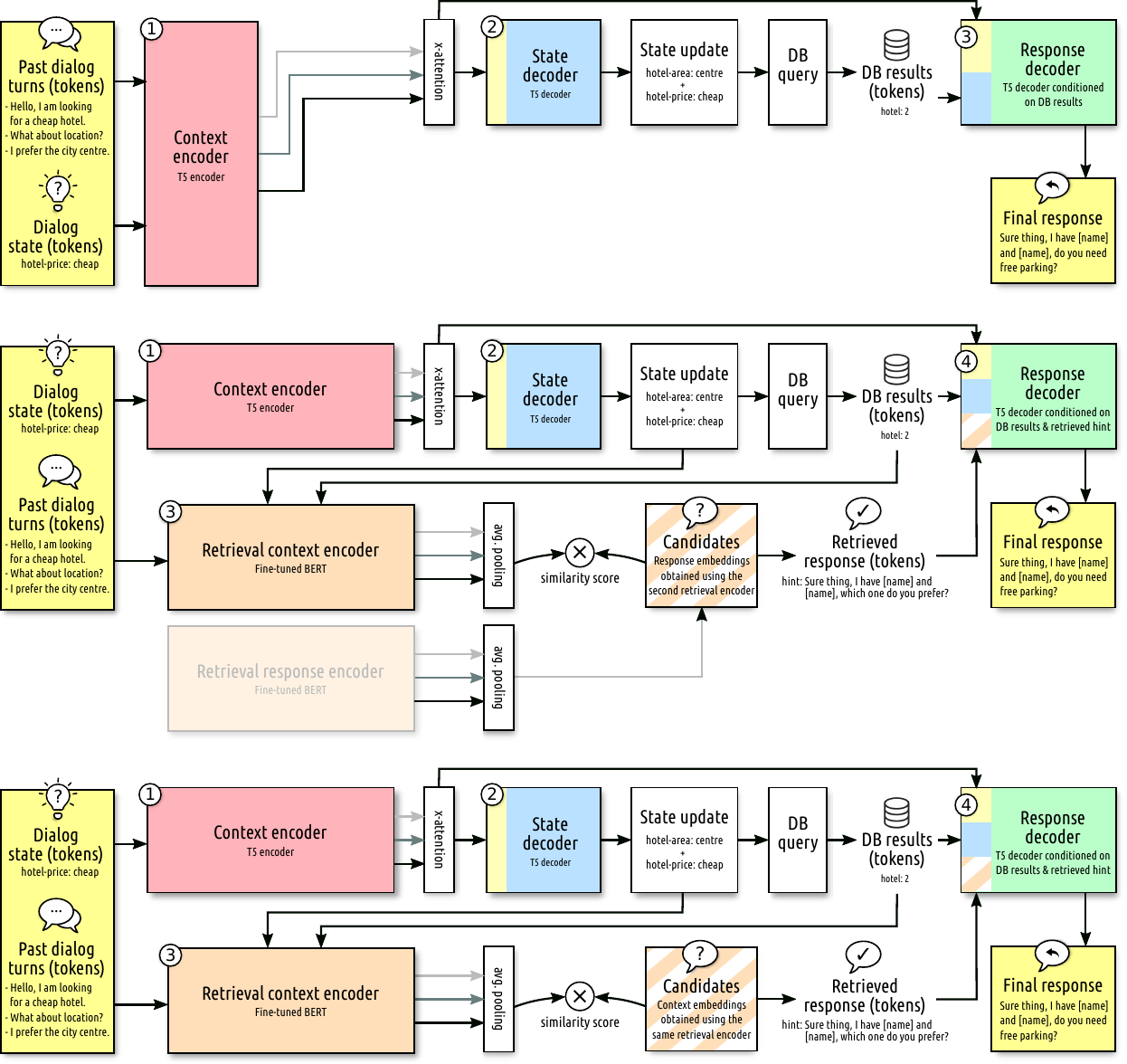}
\caption{Architecture of the baseline (\base, \emph{top}), dual-encoder-based model (\dual, \emph{middle}) and single-encoder action-aware model (\act, \emph{bottom}). Numbers in module boxes mark the order of processing during inference.}
\label{fig:pipeline_others}
\end{figure*}

Figure~\ref{fig:pipeline_others} shows architectures of the baseline (\base), dual-encoder-based model (\dual), and single-encoder action-aware model (\act). See Figure~\ref{fig:pipeline} for details about \comb and Section~\ref{sec:method} for description of the models. 


\section{Beam Search Results}
\label{sec:appendix_b}

See Table~\ref{tab:automatic_eval_beam} for the results of beam search-based response generation evaluation, and compare the results with greedy decoding evaluation (see Section~\ref{sec:response_results} and Table~\ref{tab:automatic_eval_greedy}). For all models, we used beams of size 8 during the decoding

In the case of conservative $\alpha$-blending, beam search decoding results in higher lexical diversity for all retrieval-augmented systems. However, the gains with respect to Inform and Success rates are mostly very small or not present at all in the case of \adual and \comb. All BLEU scores are slightly lower which corresponds with the higher output diversity. We notice that the numbers for the baseline without a retrieval component have an opposite trend. Beam search decoding causes lower lexical diversity and higher BLEU. We attribute this to the fact that beam search decoding prefers safer responses with a higher overall probability.

When using higher $\alpha$-blending, the differences become small even in the case of lexical diversity. We hypothesize that all the retrieval-based models are not substantially influenced by the particular response decoding strategy because they strongly rely on the retrieved hints and their copying. 

\begin{table*}[!th]
    \centering\small
     \renewcommand{\arraystretch}{0.95}
    \begin{tabular}{cl|ccccc|cc}
      \toprule
      \multicolumn{2}{c|}{Setting} & BLEU & Inform & Success & Num. trigrams & Bi-gram entropy & Hint-BLEU & Hint-copy \\
      \midrule
    
      \multicolumn{2}{c|}{\base} & $19.1 \pm 0.3$ & $73.1 \pm 1.8$ & $63.0 \pm 1.7$ & 2683  & 1.81 & - & - \\
      \midrule
      
      \parbox[t]{2mm}{\multirow{4}{*}{\rotatebox[origin=c]{90}{\emph{$\alpha=0.05$}}}} 
        &  \dual & $16.1 \pm 0.3$ & $81.1 \pm 0.5$ & $68.3 \pm 0.8$ & 10098 & \textbf{2.49} & $41.9$ & 25.2 \% \\
        & \adual & $16.0 \pm 0.4$ & $78.0 \pm 1.1$ & $65.9 \pm 1.0$ & 7378 & 2.33 & $32.6$ & 19.2 \% \\
        &  \poly & $15.9 \pm 0.4$ & $80.6 \pm 0.9$ & $66.9 \pm 1.0$ & 9470  & 2.48 & $40.7$ & 24.4 \% \\
        &  \act  & $\mathbf{16.4} \pm 0.4$ & $\mathbf{82.5} \pm 0.8$ & $\mathbf{69.8} \pm 0.6$ & \textbf{10457} & 2.46 & $44.2$ & 29.0 \% \\
        &  \comb & $16.2 \pm 0.3$ & $79.5 \pm 0.5$ & $68.0 \pm 0.3$ & 9072  & 2.36 & $36.2$ & 22.2 \%  \\
      \midrule
      
      \parbox[t]{2mm}{\multirow{4}{*}{\rotatebox[origin=c]{90}{\emph{$\alpha=0.4$}}}} 
        &  \dual & $12.3 \pm 0.3$ & $87.7 \pm 0.3$ & $68.7 \pm 0.6$ & 19103 & \textbf{3.28} & $83.1$ & 79.6 \% \\
        & \adual & $13.7 \pm 0.4$ & $77.8 \pm 0.9$ & $63.2 \pm 0.4$ & 10997 & \textbf{2.76} & $55.8$ & 50.2 \% \\
        &  \poly & $\mathbf{12.7} \pm 0.3$ & $85.9 \pm 0.9$ & $66.3 \pm 0.7$ & 17178 & 3.19 & $78.3$ & 74.2 \% \\
        &  \act  & $12.0 \pm 0.2$ & $\mathbf{90.3} \pm 0.5$ & $\mathbf{71.2} \pm 0.4$ & 19448 & 3.21 & $92.6$ & 90.1 \% \\
        &  \comb & $12.2 \pm 0.2$ & $89.3 \pm 0.3$ & $70.0 \pm 0.6$ & \textbf{19763} & 3.22 & $88.6$ & 86.0 \% \\
      
      \bottomrule
    \end{tabular}
    \caption{Beam search-based response generation on MultiWOZ using automatic metrics. For each model setup, We use beams of size 8 during response decoding and report results averaged over 8 random seeds. We compare the baseline (\base) and retrieval-based generative models (See Section~\ref{sec:method} and \ref{sec:experimental_setup}). See Section \ref{sec:automatic} for details about the metrics. Cf.~Table~\ref{tab:automatic_eval_greedy} showing results obtained using greedy decoding.}
    \label{tab:automatic_eval_beam}
\end{table*}

\section{End-to-end Conversation}
\label{sec:appendix_c}

\begin{figure*}[!t]
\centering
\includegraphics[width=1.0\textwidth]{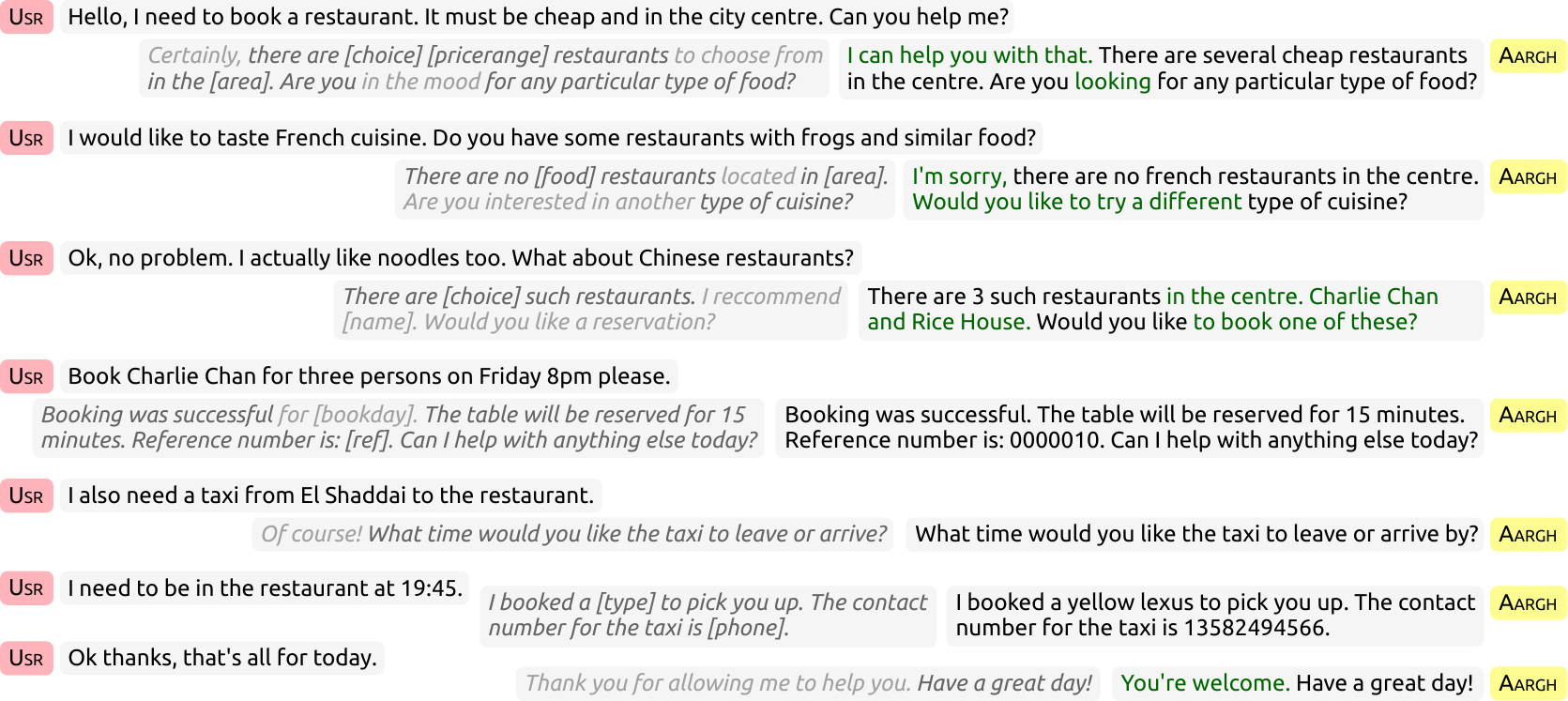}
\caption{End-to-end conversation between\definecolor{r}{RGB}{255,179,186} \tikz\draw[r, fill=r] (0,0) circle (.7ex); the user and\definecolor{yy}{RGB}{252,252,144} \tikz\draw[yy, fill=yy] (0,0) circle (.7ex); our retrieval-based \comb model with conservative $\alpha$-blending (see Section~\ref{sec:method}). For the system turns, we show delexicalized hints proposed by the retrieval module (left boxes in italics) and the corresponding lexicalized final responses (right boxes). We highlighted\definecolor{kk}{RGB}{130,130,130} \tikz\draw[kk, fill=kk] (0,0) circle (.7ex); the parts of hints present in the final texts and\definecolor{gg}{RGB}{0,130,0} \tikz\draw[gg, fill=gg] (0,0) circle (.7ex); the parts of final responses newly-introduced by the model during refining. }
\label{fig:conversation}
\end{figure*}

Figure~\ref{fig:conversation} shows a multi-domain (restaurant and taxi) end-to-end conversation between our end-to-end retrieval-based model \comb (See Section~\ref{sec:hybrid}).

\section{Human Evaluation Interface}
\label{sec:appendix_d}

\begin{figure*}[!th]
\centering
\includegraphics[width=1.0\textwidth]{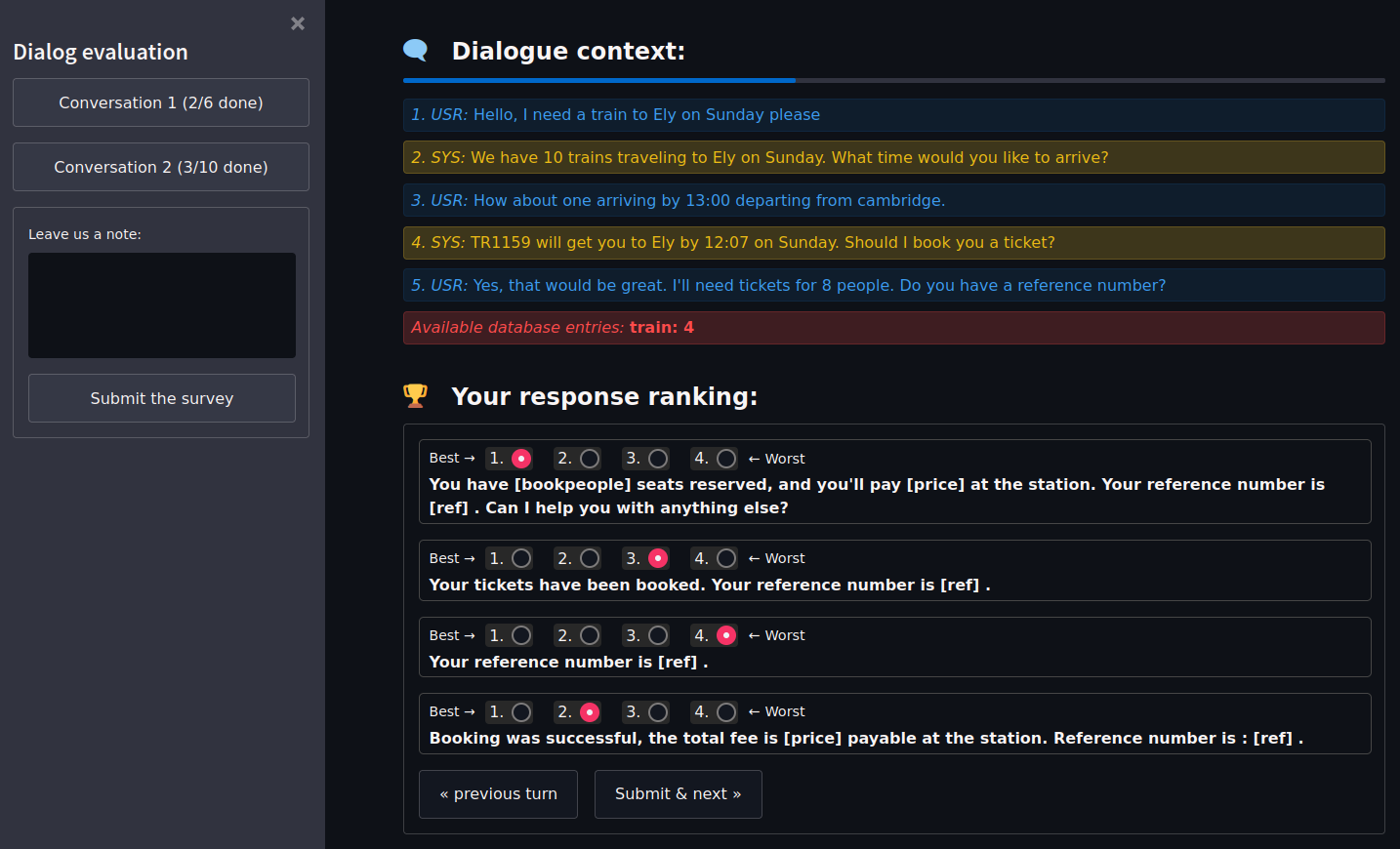}
\caption{Our graphical user interface used for human evaluation.}
\label{fig:gui}
\end{figure*}

We used the graphical user interface depicted in Figure~\ref{fig:gui} for human evaluation. A full dialog context, i.e., all past utterances corresponding to the particular turn, and the number of database results were shown to participants. We asked participants to rank provided responses from the best to the worst. They evaluated only two conversations in a single run and we sampled the conversations from the test set so that all participants receive roughly the same number of turns to assess. Evaluated responses were shown side-by-side; each of them had a dedicated discrete scale from 1 to 4 where 1 was labeled as the best and 4 as the worst. More responses could receive the same ranking. Participants could move forward and backward in the conversations and they could switch to another conversation anytime.  

\end{document}